\documentclass[journal]{IEEEtran}

\usepackage[utf8]{inputenc}
\usepackage{color}
\usepackage{xcolor}
\usepackage{array}
\usepackage{verbatim}
\usepackage{float}
\usepackage{amsmath}
\usepackage{amsthm}
\usepackage{amssymb}
\usepackage{graphicx}
\usepackage{longtable}
\usepackage{multirow}
\usepackage{booktabs}
\usepackage{balance}
\providecommand{\tabularnewline}{\\}
\usepackage{algorithmic}
\usepackage{longtable}

\floatstyle{ruled}
\newfloat{algorithm}{tbp}{loa}
\providecommand{\algorithmname}{Algorithm}
\floatname{algorithm}{\protect\algorithmname}
\usepackage{multirow}
\usepackage[unicode=true,
bookmarks=false,
breaklinks=false,pdfborder={0 0 1},colorlinks=false]
{hyperref}
\hypersetup{
	colorlinks,bookmarksopen,bookmarksnumbered,citecolor=blue,urlcolor=blue}
\usepackage{cite}

\usepackage{lipsum}
\usepackage{mathtools}
\usepackage{cuted}

\floatstyle{ruled}
\newfloat{algorithm}{tbp}{loa}
\providecommand{\algorithmname}{Algorithm}
\floatname{algorithm}{\protect\algorithmname}

\makeatletter
\let\oldforeign@language\foreign@language
\DeclareRobustCommand{\foreign@language}[1]{%
	\lowercase{\oldforeign@language{#1}}}

\DeclareTextSymbolDefault{\textquotedbl}{T1}

\let\oldforeign@language\foreign@language
\DeclareRobustCommand{\foreign@language}[1]{%
	\lowercase{\oldforeign@language{#1}}}

\ifCLASSINFOpdf
\else
\fi

\@ifundefined{showcaptionsetup}{}{%
	\PassOptionsToPackage{caption=false}{subfig}}
\usepackage{subfig}

\newcommand{\MYfooter}{\smash{
		\hfil\parbox[t][\height][t]{\textwidth}{\centering
			\thepage}\hfil\hbox{}}}

\def\ps@IEEEtitlepagestyle{%
	\def\@oddhead{\parbox[t][\height][t]{\textwidth}{\centering \scriptsize
			Personal use of this material is permitted. Permission from the author(s) and/or copyright holder(s), must be obtained for all other uses. Please contact us and provide details if you believe this document breaches copyrights.\\
			\noindent\makebox[\linewidth]{}
		}\hfil\hbox{}}%
	\def\@evenhead{\scriptsize\thepage \hfil \leftmark\mbox{}}%
	\def\@oddfoot{\parbox[t][\height][l]{\textwidth}{
			\vspace{-20pt}{\rule{\textwidth}{0.4pt}}\\ \footnotesize\underline{To cite this article:}
			{\bf{\footnotesize\textcolor{red}{A. V. Jonnalagadd, H. A. Hashim, and A. Harris, "Comprehensive and Comparative Analysis between Transfer Learning and Custom Built VGG and CNN-SVM Models for Wildfire Detection," In Proc. of the 2024 IEEE International Conference On Intelligent Computing in Data Sciences, Marrakech, Morocco, pp. 1-7, 2024.}}} \\
			\noindent\makebox[\linewidth]
		}\hfil\hbox{}}%
	\def\@evenfoot{\MYfooter}}

\makeatother
\begin{document}
	\bstctlcite{IEEEexample:BSTcontrol}

\title{Comprehensive and Comparative Analysis between Transfer Learning and Custom Built VGG and CNN-SVM Models for Wildfire Detection}

	\author{Aditya V. Jonnalagadda, Hashim A. Hashim, and Andrew Harris
		\thanks{This work was supported in part by National Sciences and Engineering
			Research Council of Canada (NSERC), under the grant number RGPIN-2022-04937.}
		\thanks{A. V. Jonnalagadda, H. A. Hashim, A. Harris are with the Department
			of Mechanical and Aerospace Engineering, Carleton University, Ottawa,
			ON, K1S-5B6, Canada (e-mail: hhashim@carleton.ca).}
	}

	
	
	\maketitle
\begin{abstract}
	Contemporary Artificial Intelligence (AI) and Machine Learning (ML)
	research places a significant emphasis on transfer learning, showcasing
	its transformative potential in enhancing model performance across
	diverse domains. This paper examines the efficiency and effectiveness
	of transfer learning in the context of wildfire detection. Three purpose-built
	models -- Visual Geometry Group (VGG)-7, VGG-10, and Convolutional
	Neural Network (CNN)-Support Vector Machine(SVM) CNN-SVM -- are rigorously
	compared with three pretrained models -- VGG-16, VGG-19, and Residual
	Neural Network (ResNet) ResNet101. We trained and evaluated these
	models using a dataset that captures the complexities of wildfires,
	incorporating variables such as varying lighting conditions, time
	of day, and diverse terrains. The objective is to discern how transfer
	learning performs against models trained from scratch in addressing
	the intricacies of the wildfire detection problem. By assessing the
	performance metrics, including accuracy, precision, recall, and F1
	score, a comprehensive understanding of the advantages and disadvantages
	of transfer learning in this specific domain is obtained. This study
	contributes valuable insights to the ongoing discourse, guiding future
	directions in AI and ML research.
\end{abstract}

\section{Introduction}
	\IEEEPARstart{T}{ransfer} learning has emerged as a focal point in contemporary Artificial
	Intelligence (AI) and Machine Learning (ML) research due to its transformative
	impact on model performance across diverse domains \cite{jonnalagadda2024segnet}.
	The collective efforts of leading companies and research institutions
	underscore the pivotal role transfer learning plays in enhancing efficiency
	across various facets of human life. This surge in popularity can
	be attributed to the adaptability and efficiency transfer learning
	imparts to AI and ML applications. In the context of AI/ML, transfer
	learning refers to leveraging pretrained models initially trained
	on a source dataset and applying them to a target dataset with shared
	domain characteristics \cite{Ref4}. Notably, in computer vision \cite{jonnalagadda2024segnet},
	models are often trained on various iterations of the ImageNet dataset
	\cite{deng2009imagenet}, which comprises an extensive array of object
	classes, including animals, everyday objects, and specific species
	of flora and fauna. This dataset continually evolves, incorporating
	new classes and images to enhance its comprehensiveness.
	
	Researchers have dedicated significant efforts to developing models
	capable of accurately classifying objects within the ImageNet dataset,
	achieving remarkable accuracies often exceeding $99\%$ \cite{Ref3}.
	For professionals engaged in specific classification tasks, such as
	identifying microorganisms or various car models, transfer learning
	offers a notable advantage. Rather than undertaking the laborious
	process of training a model from scratch, practitioners can access
	pretrained neural network weights and adapt them to specific domain-related
	tasks. Transfer learning presents broad applicability across diverse
	fields of study, including computer vision, Large Language Models
	(LLMs), language translation, and chatbots.
	
	\begin{figure*}
		\centering{}\includegraphics[scale=0.66]{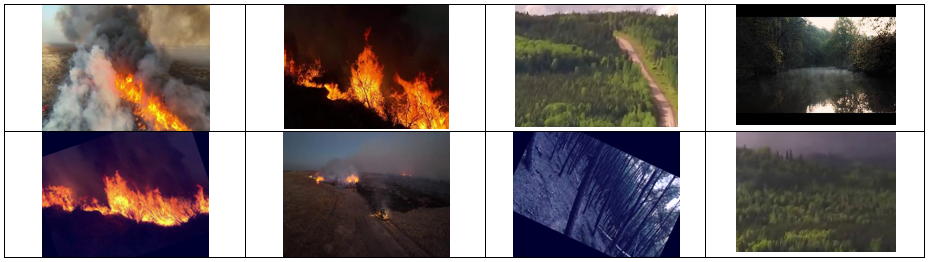}\caption{Sample of images in the training dataset. These images include synthetic
			images created by data augmentation techniques like rotation, translation,
			scaling, brightness adjustment, and the introduction of Gaussian noise
			(for more details visit \cite{jonnalagadda2024segnet}).}
		\label{fig:Fig1_Conf}
	\end{figure*}

	\subsection{Computer Vision and Language Models}
	
	\textit{Computer Vision}: Transfer learning significantly impacts
	computer vision by leveraging pretrained models on extensive datasets
	such as ImageNet. This approach expedites the learning process, enhancing
	efficiency in tasks such as object detection, image segmentation,
	and facial recognition \cite{jonnalagadda2024segnet}. The ability
	to transfer knowledge from one visual domain to another enables models
	to discern intricate patterns and features with remarkable accuracy.
	\textit{Large Language Models (LLMs)}: In natural language processing
	and LLMs, transfer learning has revolutionized the landscape. Models
	like Bidirectional Encoder Representations from Transformers (BERT)
	are pretrained on vast corpora, enabling them to grasp intricate language
	nuances and contextual relationships. These pretrained language models
	serve as a foundation for diverse Natural Language Processing (NLP)
	tasks, including sentiment analysis, text summarization, and question
	answering. \textit{Language Translation}: Transfer learning proves
	instrumental in language translation tasks. By pre-training models
	on multilingual datasets, these models gain an understanding of language
	structures and semantics across different languages. Fine-tuning for
	specific language pairs leads to more accurate and contextually relevant
	translations, contributing to the development of more accessible and
	effective communication tools. \textit{Chatbots}: The development
	of intelligent chatbots benefits significantly from transfer learning.
	Pretrained models, equipped with extensive linguistic knowledge, comprehend
	user queries, understand context, and generate coherent responses.
	This approach mitigates the need for exhaustive training on specific
	dialog datasets, allowing developers to create chatbots adept at natural
	language understanding and generation.
	
	\subsection{Pretrained Models vs Custom Built Models}
	
	Utilizing pretrained models offers substantial benefits in terms of
	data efficiency, as these models are initially trained on extensive
	and varied datasets, enabling them to grasp general features and patterns.
	This proves advantageous when dealing with limited labeled data for
	a particular task, as the pretrained model has already acquired useful
	representations from a broader context. Transfer learning further
	streamlines the process by taking a pretrained model and fine-tuning
	it on a task-specific dataset, leading to faster convergence and requiring
	less labeled data compared to training from scratch. The efficiency
	gains extend to computation time and resources, as training large
	neural networks from the beginning is computationally intensive, whereas
	pretrained models save time by having completed a significant portion
	of the learning process. Additionally, pretrained models serve as
	effective feature extractors, particularly leveraging lower-level
	features that are transferable across different tasks. The models'
	ability to generalize well to diverse inputs, especially when the
	pre-training dataset aligns with the target domain, contributes to
	improved performance on new, unseen data. Furthermore, in scenarios
	involving related but not identical domains, pretrained models support
	domain adaptation through fine-tuning, enabling effective performance
	on specific tasks within the target domain. The collaborative nature
	of the research community enhances accessibility, with many open-source
	pretrained models in computer vision available for practitioners,
	fostering the development and utilization of state-of-the-art models
	for diverse applications.
	
	\subsection{Contributions}
	
	This paper applies and studies the performance measure of transfer
	learning in the context of wildfire detection. To assess the performance
	measure of transfer learning utilization, three custom-built models,
	namely, Visual Geometry Group (VGG) VGG-7, VGG-10, and Convolutional
	Neural Network (CNN)-Support Vector Machine(SVM) CNN-SVM are compared
	relative to three pretrained models -- VGG-16, VGG-19, and Residual
	Neural Network (ResNet) ResNet101. The proposed study shows that transfer
	learning can successfully capture the complexities of wildfires, incorporating
	challenging variables (e.g., time of day, varying lighting conditions,
	and diverse terrains) as presented in Fig. \ref{fig:Fig1_Conf}. In
	this paper, we show how transfer learning can address the intricacies
	of the wildfire detection when compared to custom-built models trained
	from scratch. Performance metrics, such as accuracy, precision, recall,
	and F1 score, are used to provide a comprehensive understanding of
	transfer learning advantages and disadvantages in the wildfire detection
	domain.
	
	\subsection{Structure}
	
	The rest of the paper is organized as follows: Section \ref{sec:Problem-Formulation}
	describes the problem formulation and dataset preprocessing. Section
	\ref{sec:Methodology} illustrates the research methodology and segmentation.
	Section \ref{sec:Methodology} presents custom built CNN-SVM, VGG-7,
	and VGG-10, and pretrained VGG-16, VGG-19, and ResNet101. Section
	\ref{sec:Results} illustrates the comparative results, performance
	analysis, and test cases. Finally, Section \ref{sec:Conclusion} concludes
	the work.
	
	\section{Problem Formulation\label{sec:Problem-Formulation}}
	
	\begin{figure}
		\centering{}\subfloat[]{\includegraphics[scale=0.55]{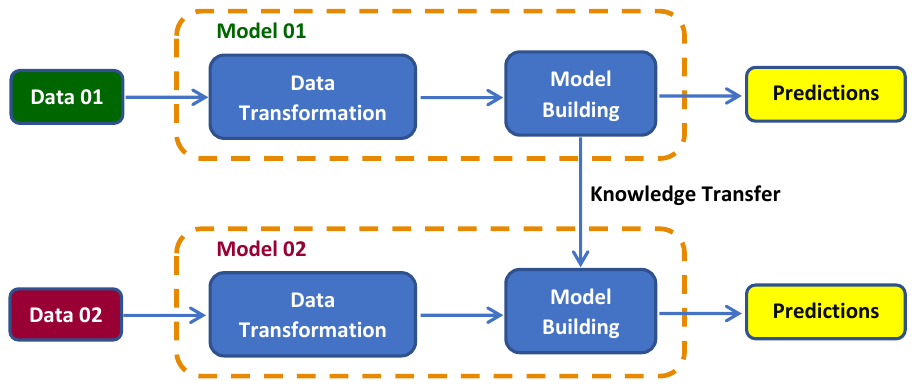}}
		
		\subfloat[]{\includegraphics[scale=0.7]{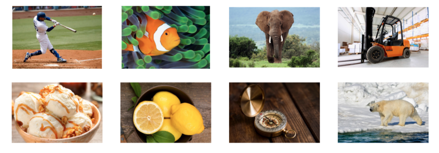}}
		
		\caption{(a) Working principle (schematic) of transfer learning and (b) Sample
			Images from ImageNet dataset \cite{deng2009imagenet}.}
		\label{fig:Fig2_Conf}
	\end{figure}
	\vspace{-0.2cm}
	
	Wildfires pose as a significant natural calamity, capable of causing
	widespread devastation if not promptly detected and addressed during
	their early stages \cite{Ref5}. Originating from seemingly innocuous
	sources like a dry tree in the forest, these fires can rapidly escalate,
	engulfing natural habitats and leading to casualties, both human and
	animal \cite{jonnalagadda2024segnet}. Recognizing the urgency of
	early detection, research in wildfire detection has gained paramount
	importance, driven by the advancements in AI and ML techniques. Drone
	is important player in our daily life applications, in particular
	vision-based activites \cite{hashim2021geometricSLAM,hashim2024uwb,hashim2023exponentially,ali2024mpc,Arezo2024Quat}.
	Among the various methods, drone surveillance has emerged as a popular
	and effective approach with advanced perception \cite{hashim2024uwb,hashim2021geometricSLAM,hashim2021geometric}
	and perception-based control algorithms \cite{hashim2023exponentially,hashim2023observer}
	with an overall objective of following regulated safety and regulation
	protocols \cite{hashim2024Avionics, hashim2024EW}. Several well-established models
	for image classification, including AlexNet, InceptionNet, LeNet,
	and ResNet, have demonstrated efficacy in the realm of wildfire detection.
	However, a critical consideration is that these models are pretrained
	on datasets like ImageNet, which do not inherently include wildfire
	instances. Consequently, leveraging these pretrained models for wildfire
	detection necessitates fine-tuning them for the specific characteristics
	of the wildfire domain and further training on a customized fire/no-fire
	dataset. This process is crucial to ensure the models' adaptability
	and accuracy in identifying wildfire-related features.
	
	The alternative approach involves training models from scratch, omitting
	the advantages offered by transfer learning (see the schematic representation
	in Fig. \ref{fig:Fig2_Conf}.(a)). Transfer learning entails taking
	models pretrained on a broad dataset and adapting them to a specific
	task. In the context of wildfire detection, the focus is on comparing
	the results obtained by building a model from scratch against a model
	with pretrained weights \cite{Ref17,Ref18}. In this paper, we compare
	VGG pretrained network and VGG custom network. This comparative analysis
	aims to shed light on the performance disparities between the two
	methodologies. The study delves into the intricacies of wildfire detection,
	considering the nuances associated with building a model from scratch
	and the benefits derived from leveraging pretrained weights. The evaluation
	encompasses factors such as detection accuracy, computational efficiency,
	and the overall effectiveness of the models in real-world scenarios.
	By comprehensively comparing these two approaches, the research aims
	to provide valuable insights that can inform decision-making in wildfire
	detection systems.
	
	Additionally, the paper explores the domain dependencies between the
	target dataset, specific to wildfire instances, and the source dataset
	used for pre-training the models \cite{Ref11}. Understanding these
	dependencies is vital for optimizing model performance, as the characteristics
	of the source dataset may influence the model's ability to discern
	relevant features in the target domain. This paper contributes to
	the ongoing discourse on wildfire detection by conducting a thorough
	comparative analysis of building models from scratch and utilizing
	pretrained weights \cite{Ref15}. By looking at the advantages and
	potential drawbacks of each approach, the study offers practical guidance
	for researchers and practitioners involved in developing effective
	wildfire detection systems. Furthermore, the investigation into domain
	dependencies enhances the understanding of the factors influencing
	model adaptation and generalization, contributing to the broader field
	of AI/ML applications in disaster management \cite{Ref16}.
	
	\subsection{Dataset and Preprocessing}
	
	Compiling an adequate wildfire image dataset presents formidable challenges
	owing to its scarcity and constrained accessibility \cite{Ref6}.
	The infrequent occurrence of wildfires, coupled with safety considerations
	and the urgent response required during active incidents, hinders
	the systematic capture of diverse and representative images. Notably,
	government restrictions on the use of real wildfire images, particularly
	those depicting environmental and human impacts, further contribute
	to the scarcity of suitable datasets. Ethical considerations surrounding
	the utilization of such images add another layer of complexity to
	dataset acquisition \cite{Ref7}. Thus, researchers and practitioners
	engaged in wildfire detection confront significant difficulties in
	assembling a sufficiently extensive and varied dataset. The limited
	availability of images, safety constraints, and regulatory restrictions
	collectively impede the comprehensive development and training of
	machine learning models essential for effective wildfire detection
	and management. Thus, in this paper a custom dataset has been created
	consisting of a mixture of individual images downloaded from the internet
	and the FLAME dataset available to download from IEEE data port. For
	this custom dataset two classes namely fire and non-fire have been
	defined. Fig. \ref{fig:Fig1_Conf} shows a few samples from both the
	classes of the custom dataset.
	
	The ImageNet dataset (Fig. \ref{fig:Fig2_Conf}.(b)) stands as a monumental
	resource in the realm of computer vision, comprising an extensive
	and diverse collection of labeled images. Created by researchers at
	Princeton University, ImageNet encompasses millions of high-resolution
	images, covering a vast array of object categories. Each image in
	the dataset is meticulously annotated, providing invaluable ground
	truth labels for various objects, animals, and scenes. ImageNet has
	been a catalyst for significant advancements in image classification
	and object recognition, serving as a benchmark for evaluating the
	performance of machine learning models \cite{Ref19}.
	
	ImageNet's influence is notably attributed to its annual Large Scale
	Visual Recognition Challenge (ILSVRC), a competition that has spurred
	the development of state-of-the-art deep learning models. Pioneering
	models like AlexNet, VGGNet, InceptionNet, and ResNet have been trained
	and benchmarked on ImageNet, showcasing their efficacy in classifying
	and recognizing objects within diverse visual contexts. The ImageNet
	dataset aids in generalizing models for wildfire detection by leveraging
	pretrained models on diverse objects \cite{Ref14}. Although ImageNet
	lacks wildfire images, the learned features from generic objects can
	be harnessed. Using transfer learning, a pretrained model's lower
	layers capture universal features, serving as a foundation. Fine-tuning
	on a custom wildfire dataset tailor the model to recognize fire-related
	patterns. This approach capitalizes on the model's prior knowledge
	from ImageNet, enhancing its adaptability and generalization to identify
	wildfires, even when specific wildfire images are scarce during initial
	training. 
	
	\section{Methodology\label{sec:Methodology}}
	
	This study contrasts non-trained models developed from scratch, specifically
	tailored for diverse landscapes such as jungles, mountains, and arid
	regions \cite{Ref9,Ref12}. The three newly crafted models include
	CNN-SVM, VGG-7, and VGG-10. In comparison, pretrained models sourced
	from the ImageNet dataset, namely VGG-16 \cite{Ref20}, VGG-19 \cite{Ref20},
	and ResNet \cite{Ref21}, are examined.
	
	\subsection{Custom Built CNN-SVM}
	
	The CNN-SVM model combines CNNs and SVMs for image classification.
	CNNs \cite{jonnalagadda2024segnet,chamberland2023autoencoder} extract
	hierarchical features, and SVMs act as a classifier. Table \ref{tab:ComprehensiveConv}
	illustrates the model architecture. This specific configuration, revealed
	in extensive testing of diverse setups, was chosen due to its optimal
	results across test, train, and validation accuracies. The model's
	architecture was meticulously tailored to align with the characteristics
	of the present dataset, ensuring a robust and effective model addressing
	the complexities inherent in the data.
	
	\subsection{Custom Built VGG-7 and VGG-10}
	
	VGG-7 and VGG-10 are custom made models built with a structure similar
	to VGG-16 and VGG-19. These models are gained by dropping pooling
	layers and convolution layers with a high number of filters. VGG-16
	is a complex deep neural network that aims at recognizing massive
	amounts of features pertaining to $1,000$ different classes in the
	ImageNet dataset. It is not designed to be trained from ground zero
	on a considerably small dataset with only two classes, in this case
	- fire and non-fire. Thus, after repetitive trials, it proved to be
	ineffective while training the VGG-16 architecture without pretrained
	weights. Another drawback to training a complex architecture like
	VGG-16 and VGG-19 from scratch is the high amount of time and memory
	power required. Hence, models like VGG-7 and VGG-10 were customized
	to fit the requirements of wildfire detection and for comparison purposes.
	Their architecture schematics are presented in Table 2 And Table 4
	for VGG-7 and VGG-10, respectively.
	
	\subsection{Pretrained VGG-16 and VGG-19}
	
	VGG-16 and VGG-19, being pretrained models on the ImageNet dataset,
	serve as a foundation for further training through transfer learning
	\cite{Ref10}. The weights, obtained from the Keras application library,
	are initially optimized for a distinct domain. In this study, all
	layers, except the top layers (input and output), are set to be untrainable
	to preserve the pre-existing knowledge. To streamline and expedite
	the learning process, a modification is introduced by consolidating
	two dense layers into a single dense layer, enhancing efficiency.
	Given the optimized nature of the weights, this adjustment aims at
	producing faster and more effective learning while adapting to the
	nuances of wildfire detection. The output layer undergoes a transformation,
	transitioning from 512 dense nodes to a singular node. This alteration
	necessitates a change in the activation function from \textit{softmax}
	to linear, aligning with the specific requirements of the binary classification
	task involving fire and non-fire instances. By leveraging the pretrained
	weights and optimizing the network architecture, this tailored approach
	ensures the transferability of knowledge from ImageNet to the targeted
	wildfire detection domain. The reason for these modifications lies
	in achieving a balance between computational efficiency and the model's
	capacity to grasp the distinctive features relevant to the task, thereby
	maximizing the benefits of transfer learning for enhanced wildfire
	detection capabilities \cite{Ref13}. The architecture is demonstrated
	by table 5 and table 6 for VGG-16 and VGG-19, respectively.
	
	\subsection{Pretrained ResNet}
	
	ResNet, or Residual Neural Network, is a deep learning architecture
	renowned for addressing vanishing gradient issues in very deep networks.
	ResNet \cite{Ref21} employs residual blocks, allowing direct paths
	for information flow and facilitating the training of extremely deep
	neural networks. ResNet101, an extension of ResNet, employs 101 layers
	in its deep neural network architecture. It introduces residual blocks
	with skip connections, addressing vanishing gradient issues and facilitating
	training of extremely deep networks. The model's depth enables capturing
	intricate features, enhancing its performance in complex visual recognition
	tasks. For this paper, this model has been modified similar to the
	other pretrained VGG models, with its output layer modified to converge
	from 512 nodes to a single node for this binary classification task.
	
	\section{Results\label{sec:Results}}
	
	\begin{table*}
		\caption{\label{tab:ComprehensiveConv}Proposed VGG-7, CNN-SVM, VGG-10, VGG-16,
			and VGG-19 architecture.}
		
		\centering{}%
		\begin{tabular}{>{\raggedright}m{2.8cm}l>{\centering}p{1.9cm}>{\centering}p{1.9cm}>{\centering}p{1.9cm}>{\centering}p{1.9cm}>{\centering}p{1.9cm}}
			\hline 
			\multicolumn{1}{l}{} &  & VGG-7 & CNN-SVM  & VGG-10  & VGG-16 & VGG-19 \tabularnewline
			\hline 
			\multirow{5}{2.8cm}{Convolution Layer 1} & Padding: & SAME & SAME & SAME & SAME & SAME\tabularnewline
			\cline{2-7} \cline{3-7} \cline{4-7} \cline{5-7} \cline{6-7} \cline{7-7} 
			& Activation: & ReLu & ReLu & ReLu & ReLu & ReLu\tabularnewline
			\cline{2-7} \cline{3-7} \cline{4-7} \cline{5-7} \cline{6-7} \cline{7-7} 
			& Filters: & 64 & 32 & 64 & 64 & 64\tabularnewline
			\cline{2-7} \cline{3-7} \cline{4-7} \cline{5-7} \cline{6-7} \cline{7-7} 
			& Kernel: & 3 & 3 & 3 & 3 & 3\tabularnewline
			\cline{2-7} \cline{3-7} \cline{4-7} \cline{5-7} \cline{6-7} \cline{7-7} 
			& Number of Layers: & 2 & 1 & 2 & 2 & 2\tabularnewline
			\hline 
			MaxPool Layer 1  &  & 2X2 & 2X2, Stride = 2 & 2X2 & 2X2 & 2X2\tabularnewline
			\hline 
			\multirow{5}{2.8cm}{Convolution Layer 2} & Padding: & SAME & SAME & SAME & SAME & SAME\tabularnewline
			\cline{2-7} \cline{3-7} \cline{4-7} \cline{5-7} \cline{6-7} \cline{7-7} 
			& Activation: & ReLu & ReLu & ReLu & ReLu & ReLu\tabularnewline
			\cline{2-7} \cline{3-7} \cline{4-7} \cline{5-7} \cline{6-7} \cline{7-7} 
			& Filters: & 128 & 64 & 128 & 128 & 64\tabularnewline
			\cline{2-7} \cline{3-7} \cline{4-7} \cline{5-7} \cline{6-7} \cline{7-7} 
			& Kernel: & 3 & 3 & 3 & 3 & 3\tabularnewline
			\cline{2-7} \cline{3-7} \cline{4-7} \cline{5-7} \cline{6-7} \cline{7-7} 
			& Number of Layers: & 2 & 1 & 2 & 2 & 2\tabularnewline
			\hline 
			MaxPool Layer 2  &  & 2X2 & 2X2, Stride = 2 & 2X2 & 2X2 & 2X2\tabularnewline
			\hline 
			\multirow{5}{2.8cm}{Convolution Layer 3} & Padding: & - & SAME & SAME & SAME & SAME\tabularnewline
			\cline{2-7} \cline{3-7} \cline{4-7} \cline{5-7} \cline{6-7} \cline{7-7} 
			& Activation: & - & ReLu & ReLu & ReLu & ReLu\tabularnewline
			\cline{2-7} \cline{3-7} \cline{4-7} \cline{5-7} \cline{6-7} \cline{7-7} 
			& Filters: & - & 128 & 256 & 256 & 256\tabularnewline
			\cline{2-7} \cline{3-7} \cline{4-7} \cline{5-7} \cline{6-7} \cline{7-7} 
			& Kernel: & - & 3 & 3 & 3 & 3\tabularnewline
			\cline{2-7} \cline{3-7} \cline{4-7} \cline{5-7} \cline{6-7} \cline{7-7} 
			& Number of Layers: & - & 1 & 3 & 3 & 4\tabularnewline
			\hline 
			MaxPool Layer 3  &  & - & 2X2, Stride = 2 & 2X2 & 2X2 & 2X2\tabularnewline
			\hline 
			\multirow{5}{2.8cm}{Convolution Layer 4} & Padding: & - & - & - & SAME & SAME\tabularnewline
			\cline{2-7} \cline{3-7} \cline{4-7} \cline{5-7} \cline{6-7} \cline{7-7} 
			& Activation: & - & - & - & ReLu & ReLu\tabularnewline
			\cline{2-7} \cline{3-7} \cline{4-7} \cline{5-7} \cline{6-7} \cline{7-7} 
			& Filters: & - & - & - & 512 & 512\tabularnewline
			\cline{2-7} \cline{3-7} \cline{4-7} \cline{5-7} \cline{6-7} \cline{7-7} 
			& Kernel: & - & - & - & 3 & 3\tabularnewline
			\cline{2-7} \cline{3-7} \cline{4-7} \cline{5-7} \cline{6-7} \cline{7-7} 
			& Number of Layers: & - & - & - & 3 & 4\tabularnewline
			\hline 
			MaxPool Layer 4  &  & - & - & - & 2X2 & 2X2\tabularnewline
			\hline 
			\multirow{5}{2.8cm}{Convolution Layer 5} & Padding: & - & - & - & SAME & SAME\tabularnewline
			\cline{2-7} \cline{3-7} \cline{4-7} \cline{5-7} \cline{6-7} \cline{7-7} 
			& Activation: & - & - & - & ReLu & ReLu\tabularnewline
			\cline{2-7} \cline{3-7} \cline{4-7} \cline{5-7} \cline{6-7} \cline{7-7} 
			& Filters: & - & - & - & 1,024 & 1,024\tabularnewline
			\cline{2-7} \cline{3-7} \cline{4-7} \cline{5-7} \cline{6-7} \cline{7-7} 
			& Kernel: & - & - & - & 3 & 3\tabularnewline
			\cline{2-7} \cline{3-7} \cline{4-7} \cline{5-7} \cline{6-7} \cline{7-7} 
			& Number of Layers: & - & - & - & 3 & 4\tabularnewline
			\hline 
			MaxPool Layer 5  &  & - & - & - & 2X2 & 2X2\tabularnewline
			\hline 
			\noalign{\vskip\doublerulesep}
			\multicolumn{7}{c}{Flatten Layer}\tabularnewline[\doublerulesep]
			\hline 
			\multirow{2}{2.8cm}{Dense Layer 1} & Nodes: & 16 & 16 & 16 & 512 & 512\tabularnewline
			\cline{2-7} \cline{3-7} \cline{4-7} \cline{5-7} \cline{6-7} \cline{7-7} 
			& Activation: & ReLu & ReLu & ReLu & ReLu & ReLu\tabularnewline
			\hline 
			\multirow{2}{2.8cm}{Dense Layer 2} & Nodes: & 16 & - & 16 & - & -\tabularnewline
			\cline{2-7} \cline{3-7} \cline{4-7} \cline{5-7} \cline{6-7} \cline{7-7} 
			& Activation: & ReLu & - & ReLu & - & -\tabularnewline
			\hline 
			\multirow{3}{2.8cm}{Output Layer } & Nodes: & 1 & 1 & 1 & 1 & 1\tabularnewline
			\cline{2-7} \cline{3-7} \cline{4-7} \cline{5-7} \cline{6-7} \cline{7-7} 
			& Activation: & Linear & Linear & Linear & softmax & softmax\tabularnewline
			\cline{2-7} \cline{3-7} \cline{4-7} \cline{5-7} \cline{6-7} \cline{7-7} 
			& Regularizer: & - & $\mathcal{L}_{2}$ & - & - & -\tabularnewline
			\hline 
			\multirow{3}{2.8cm}{Parameters} & Total: & 10,090,865 & 2,409,569 & 6,650,993 & 14,977,857 & 20,287,553\tabularnewline
			\cline{2-7} \cline{3-7} \cline{4-7} \cline{5-7} \cline{6-7} \cline{7-7} 
			& Trainable: & 10,090,865 & 2,409,569 & 6,650,993 & 263,169 & 263,169\tabularnewline
			\cline{2-7} \cline{3-7} \cline{4-7} \cline{5-7} \cline{6-7} \cline{7-7} 
			& Non-trainable: & 0 & 0 & 0 & 14,714,688 & 20,024,384\tabularnewline
			\hline 
		\end{tabular}
	\end{table*}

	\subsection{Software, Libraries Used, and Dataset}
	
	The hardware used for building and testing these models are: GPU --
	GTX 1660 TI MaxQ design, CPU -- AMD Ryzen 9 4900HS, Software --
	Python 3.11.4, TensorFlow 2.9.1, Keras 2.9.0, Libraries - NumPy, PILLOW,
	MATPLOTLIB. Breakdown of dataset images in the training and validation
	set are as follows: Dataset training are made of 3,629 total images
	(2,080 fire images and 1,549 non-fire images) while dataset validation
	are made of 385 total images (215 fire images and 170 non-fire images).
	
	\subsection{Compartive Results}
	
	\begin{figure}
		\centering{}\includegraphics[scale=0.29]{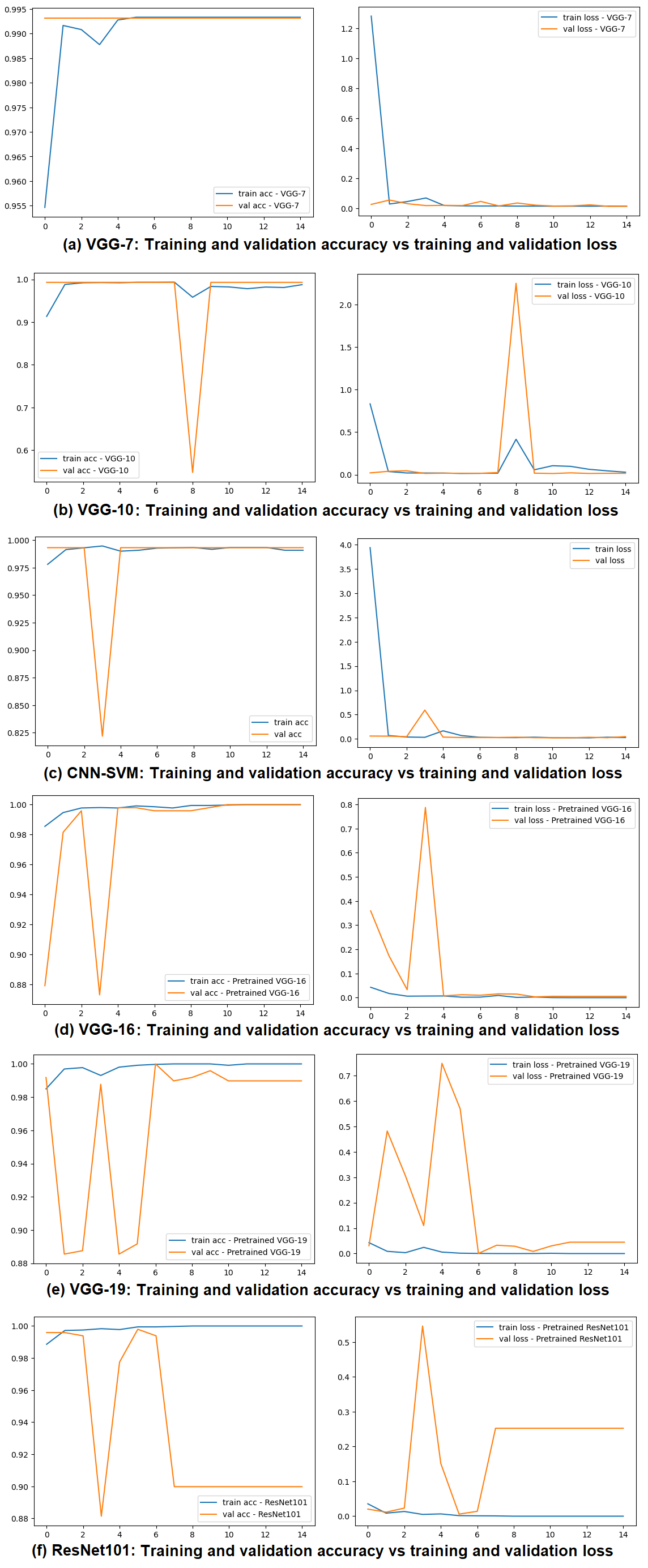}\caption{Comparative analysis and performance measures presenting training
			and validation accuracy versus training and validation loss of the
			proposed (a) VGG-7, (b) VGG-10, (c) CNN-SVM, (d) VGG-16, (e) VGG-19,
			and (f) ResNet101.}
		\label{fig:Fig3_Conf}
	\end{figure}
	
	The architectures of VGG-7, CNN-SVM, VGG-10, VGG-16, and VGG-19 is
	detailed in Table \ref{tab:ComprehensiveConv}. The pretrained ResNet101
	parameters include total parameters of 43,707,777, trainable parameters
	of 1,049,601, and non-trainable parameters of 42,658,176. As presented
	in Table \ref{tab:Performance}, the evaluation of custom models,
	specifically VGG-7 and VGG-10, alongside pretrained models, VGG-16
	and ResNet101, reveals compelling insights into their performance
	in wildfire detection. The custom VGG-7 demonstrates an impressive
	validation accuracy of $99.32\%$, coupled with a low loss of $0.0133\%$.
	Similarly, the VGG-10 model exhibits a validation accuracy of $99.30\%$,
	with a slightly higher loss of $0.0151\%$. Comparatively, the pretrained
	VGG-16 stands out with an exceptional validation accuracy of $99.99\%$,
	accompanied by a notably lower loss of $0.0056\%$. On the other hand,
	the pretrained ResNet101 exhibits a validation accuracy of $89.98\%$,
	with a comparatively higher loss of $0.2526\%$ as shown in Table
	\ref{tab:ComprehensiveConv}. The high validation accuracies achieved
	by both custom and pretrained models underscore their efficacy in
	accurately classifying wildfire instances. The custom models VGG-7
	and VGG-10, show robust performance, with accuracy exceeding $99\%$,
	indicating their suitability for the specified task. In contrast,
	the pretrained VGG-16 outperforms them all, achieving near-perfect
	accuracy, highlighting the advantages of leveraging pre-existing knowledge
	from the ImageNet dataset. The pretrained ResNet101 shows a slightly
	lower validation accuracy of $89.98\%$, suggesting that while ResNet101
	excels in various image recognition tasks, its performance may vary
	in the context of wildfire detection. These loss values provide additional
	insights into the models precision, with lower loss values indicating
	more accurate predictions.
	
	In Table \ref{tab:Performance} the values of performance metrics
	for all the chosen models for comparison is presented. These metrics
	provide an insight on how effective a model is respective to the situation
	being investigated. Various situations include, but not limited to,
	ability to detect wildfires, ability to avoid false alarms, proper
	classification of labels into respective classes of fire and non-fire,
	etc. Notably, the pretrained models, especially VGG-16, exhibit superior
	overall performance, excelling in recall, accuracy, and precision.
	VGG-16 stands out for its remarkably low false positive rate ($0.89\%$)
	and false negative rate ($0.31\%$), showcasing its effectiveness
	in minimizing both false alarms and missed wildfire instances. VGG-19
	closely follows, consistently performing well across all metrics,
	including a low false positive rate of $1.33\%$ and a false negative
	rate of $0.61\%$. In contrast, ResNet101, while delivering decent
	results, lags behind the VGG models, notably in accuracy ($92.73\%$)
	and recall ($92.61\%$). Fig. \ref{fig:Fig3_Conf} showcases the model
	evaluation graphs for all the models. Fig. \ref{fig:Fig3_Conf}.(a)
	of VGG-7 show a steady training and validation curve. Whereas, the
	other custom built models (Fig. \ref{fig:Fig3_Conf}.(b) and \ref{fig:Fig3_Conf}.(c))
	represent anomalies in their validation phase. The pretrained models
	(Fig. \ref{fig:Fig3_Conf}.(d), \ref{fig:Fig3_Conf}.(e), and \ref{fig:Fig3_Conf}.(f))
	take longer to converge to the solution, but are more accurate than
	the custom built models. These fluctuations in curves are a result
	of their prior knowledge being fine tuned to address the current problem.
	
	\subsection{Performance Analysis and Test Cases}
	
	Table \ref{tab:TruePositive} summarizes the performance of wildfire
	detection models across different architectures. Custom models VGG-7,
	VGG-10, and CNN-SVM, along with pretrained models VGG-16, VGG-19,
	and ResNet101, were evaluated against a dataset of 550 test cases,
	comprising 325 fire instances and 225 non-fire instances. The metrics
	include true positive, false positive, false negative, and true negative
	values. Notably, pretrained VGG-16 excelled with 324 true positives,
	2 false positives, 1 false negative, and 223 true negatives, showcasing
	its robust performance in accurately detecting wildfires. Turning
	our attention to the custom models, CNN-SVM emerges as a standout
	performer with the highest recall ($98.46\%$), underscoring its robust
	wildfire detection capabilities. VGG-7 and VGG-10, while demonstrating
	commendable performance, particularly in precision and accuracy, exhibit
	slightly higher false positive rates ($5.34\%$ and $4.44\%$, respectively)
	and false negative rates ($2.15\%$ and $2.46\%$, respectively) compared
	to their pretrained counterparts.
	
	\begin{table}[h]
		\caption{\label{tab:Performance}Model evaluation metrics during the training,
			testing and validation phase.}
		
		\begin{tabular}{>{\centering}p{1.1cm}>{\centering}p{0.8cm}>{\centering}p{0.8cm}>{\centering}p{0.9cm}>{\centering}p{0.8cm}>{\centering}p{0.8cm}>{\centering}p{0.8cm}}
			\cline{2-7} \cline{3-7} \cline{4-7} \cline{5-7} \cline{6-7} \cline{7-7} 
			\noalign{\vskip\doublerulesep}
			& \multicolumn{3}{c}{Custom $\%$} & \multicolumn{3}{c}{Pretrained $\%$}\tabularnewline[\doublerulesep]
			\hline 
			\noalign{\vskip\doublerulesep}
			& VGG-7 & VGG-10 & CNN-SVM & VGG-16 & VGG-19 & ResNet
			
			101\tabularnewline[\doublerulesep]
			\hline 
			\noalign{\vskip\doublerulesep}
			Training Accuracy & $99.33$ & $98.78$ & $99.1$ & $100\%$ & $100$ & $100$\tabularnewline[\doublerulesep]
			\hline 
			\noalign{\vskip\doublerulesep}
			Training Loss & $0.014$ & $0.029$ & $0.022$ & $0$ & $0$ & $0$\tabularnewline[\doublerulesep]
			\hline 
			\noalign{\vskip\doublerulesep}
			Validation Accuracy & $99.32$ & $99.3$ & $98.4$ & $99.99$ & $98.98$ & $89.98$\tabularnewline[\doublerulesep]
			\hline 
			\noalign{\vskip\doublerulesep}
			Recall & $97.84$ & $97.54$ & $98.46$ & $99.69$ & $99.38$ & $92.61$\tabularnewline[\doublerulesep]
			\hline 
			\noalign{\vskip\doublerulesep}
			Accuracy & $96.54$ & $96.72$ & $96.91$ & $99.45$ & $99.09$ & $92.73$\tabularnewline[\doublerulesep]
			\hline 
			\noalign{\vskip\doublerulesep}
			Precision & $96.36$ & $96.94$ & $96.38$ & $99.38$ & $99.08$ & $94.95$\tabularnewline[\doublerulesep]
			\hline 
		\end{tabular}
	\end{table}
	
	\begin{table}[h]
		\caption{\label{tab:TruePositive}True positives, False Positives, False Negatives
			and True Negatives predicted by the models in comparison, over a set
			of 550 total test images, with 325 images and 225 images from fire
			and non-fire classes, respectively.}
		
		\begin{tabular}{>{\raggedright}m{1.2cm}>{\centering}p{1.3cm}>{\centering}p{3.6cm}>{\centering}p{1.1cm}}
			\cline{3-4} \cline{4-4} 
			\noalign{\vskip\doublerulesep}
			&  & Total Test Cases - Fire/NonFire & True Positive\tabularnewline[\doublerulesep]
			\hline 
			\noalign{\vskip\doublerulesep}
			\multirow{3}{1.2cm}{Custom} & VGG-7 & 550 -- 325/225 & 318\tabularnewline[\doublerulesep]
			\cline{2-4} \cline{3-4} \cline{4-4} 
			\noalign{\vskip\doublerulesep}
			& VGG-10 & 550 -- 325/225 & 317\tabularnewline[\doublerulesep]
			\cline{2-4} \cline{3-4} \cline{4-4} 
			\noalign{\vskip\doublerulesep}
			& CNN-SVM & 550 -- 325/225 & 320\tabularnewline[\doublerulesep]
			\hline 
			\noalign{\vskip\doublerulesep}
			\multirow{3}{1.2cm}{Pretrained} & VGG-16 & 550 -- 325/225 & 324\tabularnewline[\doublerulesep]
			\cline{2-4} \cline{3-4} \cline{4-4} 
			\noalign{\vskip\doublerulesep}
			& VGG-19 & 550 -- 325/225 & 323\tabularnewline[\doublerulesep]
			\cline{2-4} \cline{3-4} \cline{4-4} 
			\noalign{\vskip\doublerulesep}
			& ResNet101 & 550 -- 325/225 & 301\tabularnewline[\doublerulesep]
			\hline 
		\end{tabular}
	\end{table}

	\subsection{Comments on Findings}
	
	These findings underscore the notable advantages of employing pretrained
	models, especially VGG-16 and VGG-19, which achieve optimal wildfire
	detection outcomes with minimal false positives and false negatives.
	The incorporation of false positive and false negative rates throughout
	the analysis enhances the practical relevance of our findings. For
	instance, VGG-16's exceptional performance, characterized by a false
	positive rate of $0.89\%$, indicates its ability to significantly
	reduce false alarms, crucial for minimizing unnecessary interventions.
	Simultaneously, the low false negative rate ($0.31\%$) implies a
	high sensitivity in detecting actual wildfires, essential for timely
	responses. Such insights provide valuable guidance for decision-makers,
	emphasizing the trade-offs between false positives and false negatives
	based on specific needs. This analysis contributes meaningfully to
	ongoing discussions on optimizing AI/ML applications in disaster management,
	aligning theoretical considerations with the practical implications
	of model selection in real-world wildfire scenarios.
	
	\section{Conclusion \label{sec:Conclusion}}
	
	In conclusion, the evaluation of wildfire detection models provides
	insights into their performance. Custom built models trained from
	the beginning can be very task specific. Images of wide areas of forests
	or Mediterranean landscapes, do not contain diverse objects in the
	frame in an ideal scenario. Although, objects like fall coloured trees,
	that portray an orange-red hue may confuse the model into predicting
	a false positive. Another example of a real scenario could be man-made
	objects like watch towers, fences, flag poles or flags being a part
	of the captured image in a forest, that could confuse the model not
	trained to detect these objects. Whereas, a pretrained model trained
	for wildfire instances with prior knowledge of other $1,000$ classes
	of objects could avoid these false alarms. This makes pretrained models
	to be more generalized for real-life scenarios. One major advantage
	that custom built models have over pretrained models is the computational
	speed. Pretrained models have the capability to detect thousands of
	classes due to their deep neural network. The deeper the neural network,
	higher are the number of computations required before determining
	the end result. Thus, pretrained models that contain deep neural networks
	with higher number of layers than a custom-built model, are slower
	at computational speed. This makes custom-built models more suitable
	for real-time applications.
	
	For further research, exploring hybrid models that amalgamate the
	strengths of custom and pretrained architectures could enhance detection
	capabilities. Investigating domain-specific fine-tuning for pretrained
	models and evaluating their adaptability to dynamic wildfire scenarios
	will be crucial. Additionally, delving into ensemble approaches to
	leverage diverse models for improved accuracy and reliability could
	be a promising avenue.

	\bibliographystyle{IEEEtran}
	\bibliography{bib_WildConf}

\begin{thebibliography}{10}
\providecommand{\url}[1]{#1}
\csname url@samestyle\endcsname
\providecommand{\newblock}{\relax}
\providecommand{\bibinfo}[2]{#2}
\providecommand{\BIBentrySTDinterwordspacing}{\spaceskip=0pt\relax}
\providecommand{\BIBentryALTinterwordstretchfactor}{4}
\providecommand{\BIBentryALTinterwordspacing}{\spaceskip=\fontdimen2\font plus
\BIBentryALTinterwordstretchfactor\fontdimen3\font minus
  \fontdimen4\font\relax}
\providecommand{\BIBforeignlanguage}[2]{{%
\expandafter\ifx\csname l@#1\endcsname\relax
\typeout{** WARNING: IEEEtran.bst: No hyphenation pattern has been}%
\typeout{** loaded for the language `#1'. Using the pattern for}%
\typeout{** the default language instead.}%
\else
\language=\csname l@#1\endcsname
\fi
#2}}
\providecommand{\BIBdecl}{\relax}
\BIBdecl

\bibitem{jonnalagadda2024segnet}
A.~V. Jonnalagadda and H.~A. Hashim, ``{S}eg{N}et: A segmented deep learning
  based convolutional neural network approach for drones wildfire detection,''
  \emph{Remote Sensing Applications: Society and Environment}, p. 101181, 2024.

\bibitem{Ref4}
V.~Kalaivani and P.~Chanthiya, ``A novel custom optimized convolutional neural
  network for a satellite image by using forest fire detection,'' \emph{Earth
  Science Informatics}, vol.~15, no.~2, pp. 1285--1295, 2022.

\bibitem{deng2009imagenet}
J.~Deng, W.~Dong, R.~Socher, L.-J. Li, K.~Li, and L.~Fei-Fei, ``Imagenet: A
  large-scale hierarchical image database,'' in \emph{2009 IEEE conference on
  computer vision and pattern recognition}.\hskip 1em plus 0.5em minus
  0.4em\relax Ieee, 2009, pp. 248--255.

\bibitem{Ref3}
F.~Zhuang and et~al., ``A comprehensive survey on transfer learning,''
  \emph{Proceedings of the IEEE}, vol. 109, no.~1, pp. 43--76, 2020.

\bibitem{Ref5}
S.~J. Pan and Q.~Yang, ``A survey on transfer learning,'' \emph{IEEE
  Transactions on knowledge and data engineering}, vol.~22, no.~10, pp.
  1345--1359, 2009.

\bibitem{hashim2021geometricSLAM}
H.~A. Hashim, ``A geometric nonlinear stochastic filter for simultaneous
  localization and mapping,'' \emph{Aerospace Science and Technology}, vol.
  111, p. 106569, 2021.

\bibitem{hashim2024uwb}
H.~A. Hashim, A.~E. Eltoukhy, and K.~G. Vamvoudakis, ``{UWB} ranging and {IMU}
  data fusion: Overview and nonlinear stochastic filter for inertial
  navigation,'' \emph{IEEE Transactions on Intelligent Transportation Systems},
  vol.~25, pp. 359--369, 2024.

\bibitem{hashim2023exponentially}
H.~A. Hashim, ``Exponentially stable observer-based controller for
  {VTOL}-{UAV}s without velocity measurements,'' \emph{International Journal of
  Control}, vol.~96, no.~8, pp. 1946--1960, 2023.

\bibitem{ali2024mpc}
A.~M. Ali, A.~Gupta, and H.~A. Hashim, ``Deep {R}einforcement {L}earning for
  sim-to-real policy transfer of {VTOL}-{UAV}s offshore docking operations,''
  \emph{Applied Soft Computing}, vol. 162, p. 111843, 2024.

\bibitem{Arezo2024Quat}
A.~Shevidi and H.~A. Hashim, ``Quaternion-based adaptive backstepping fast
  terminal sliding mode control for quadrotor {UAV}s with finite time
  convergence,'' \emph{Results in Engineering}, vol.~1, pp. 1--12, 2024.

\bibitem{hashim2021geometric}
H.~A. Hashim, M.~Abouheaf, and M.~A. Abido, ``Geometric stochastic filter with
  guaranteed performance for autonomous navigation based on {IMU} and feature
  sensor fusion,'' \emph{Control Engineering Practice}, vol. 116, p. 104926,
  2021.

\bibitem{hashim2023observer}
H.~A. Hashim, A.~E. Eltoukhy, and A.~Odry, ``Observer-based controller for
  {VTOL}-{UAV}s tracking using direct vision-aided inertial navigation
  measurements,'' \emph{ISA transactions}, vol. 137, pp. 133--143, 2023.

\bibitem{hashim2024Avionics}
D.~Wanner, H.~A. Hashim, S.~Srivastava, and A.~Steinhauer, ``{UAV} avionics
  safety, certification, accidents, redundancy, integrity and reliability: A
  comprehensive review and future trends,'' \emph{Drone Systems and
  Applications}, vol.~1, pp. 1--20, 2024.

\bibitem{hashim2024EW}
A.~Yu, J.~Kolotylo, H.~A. Hashim, and P.~Rennison, ``Electronic warfare
  cyberattacks, countermeasures and defensive aids of {UAV} avionics: A
  survey,'' \emph{Green Energy and Intelligent Transportation}, vol.~PP, p.~PP,
  2024.

\bibitem{Ref17}
P.~Kora and et~al., ``Transfer learning techniques for medical image analysis:
  A review,'' \emph{Biocybernetics and Biomedical Engineering}, vol.~42, no.~1,
  pp. 79--107, 2022.

\bibitem{Ref18}
M.~Jiang and et~al., ``Transfer learning-based dynamic multiobjective
  optimization algorithms,'' \emph{IEEE Transactions on Evolutionary
  Computation}, vol.~22, no.~4, pp. 501--514, 2017.

\bibitem{Ref11}
E.~Cetinic, T.~Lipic, and S.~Grgic, ``Fine-tuning convolutional neural networks
  for fine art classification,'' \emph{Expert Systems with Applications}, vol.
  114, pp. 107--118, 2018.

\bibitem{Ref15}
V.~Linardos, M.~Drakaki, P.~Tzionas, and Y.~L. Karnavas, ``Machine learning in
  disaster management: recent developments in methods and applications,''
  \emph{Machine Learning and Knowledge Extraction}, vol.~4, no.~2, 2022.

\bibitem{Ref16}
B.~Wang and et~al., ``Pre-trained language models in biomedical domain: A
  systematic survey,'' \emph{ACM Computing Surveys}, vol.~56, no.~3, pp. 1--52,
  2023.

\bibitem{Ref6}
G.~Luo and et~al., ``A thorough review of models, evaluation metrics, and
  datasets on image captioning,'' \emph{IET Image Processing}, vol.~16, no.~2,
  pp. 311--332, 2022.

\bibitem{Ref7}
A.~S. Mahdi and S.~A. Mahmood, ``Analysis of deep learning methods for early
  wildfire detection systems,'' in \emph{2022 5th International Conference on
  Engineering Technology and its Applications (IICETA)}.\hskip 1em plus 0.5em
  minus 0.4em\relax IEEE, 2022, pp. 271--276.

\bibitem{Ref19}
J.~Jiang, Y.~Shu, J.~Wang, and M.~Long, ``Transferability in deep learning: A
  survey,'' \emph{arXiv preprint arXiv:2201.05867}, 2022.

\bibitem{Ref14}
J.~Deng and et~al., ``Imagenet: A large-scale hierarchical image database,'' in
  \emph{2009 IEEE conference on computer vision and pattern recognition}.\hskip
  1em plus 0.5em minus 0.4em\relax Ieee, 2009, pp. 248--255.

\bibitem{Ref9}
C.~Picard, J.~Schiffmann, and F.~Ahmed, ``Dated: Guidelines for creating
  synthetic datasets for engineering design applications,'' \emph{arXiv
  preprint arXiv:2305.09018}, 2023.

\bibitem{Ref12}
K.~He, X.~Zhang, S.~Ren, and J.~Sun, ``Identity mappings in deep residual
  networks,'' in \emph{14th European Conference Computer Vision 2016,
  Amsterdam, Netherlands}.\hskip 1em plus 0.5em minus 0.4em\relax Springer,
  2016, pp. 630--645.

\bibitem{Ref20}
M.~A. Akhloufi, A.~Couturier, and N.~A. Castro, ``Unmanned aerial vehicles for
  wildland fires: Sensing, perception, cooperation and assistance,''
  \emph{Drones}, vol.~5, no.~1, p.~15, 2021.

\bibitem{Ref21}
K.~Simonyan and A.~Zisserman, ``Very deep convolutional networks for
  large-scale image recognition,'' \emph{arXiv preprint arXiv:1409.1556}, 2014.

\bibitem{chamberland2023autoencoder}
O.~Chamberland, M.~Reckzin, and H.~A. Hashim, ``An autoencoder with
  convolutional neural network for surface defect detection on cast
  components,'' \emph{Journal of Failure Analysis and Prevention}, vol.~23,
  no.~4, pp. 1633--1644, 2023.

\bibitem{Ref10}
D.~R. Thomas and et~al., ``Ethical issues in research using datasets of illicit
  origin,'' in \emph{Proceedings of the 2017 Internet Measurement Conference},
  2017, pp. 445--462.

\bibitem{Ref13}
A.~Krizhevsky, I.~Sutskever, and G.~E. Hinton, ``Imagenet classification with
  deep convolutional neural networks,'' \emph{Advances in neural information
  processing systems}, vol.~25, 2012.

\end{thebibliography}
\end{document}